\documentclass[10pt,twocolumn,letterpaper]{article}

\usepackage[pagenumbers]{cvpr}

\definecolor{cvprblue}{rgb}{0.21,0.49,0.74}
\usepackage[pagebackref,hypertexnames=false,breaklinks,colorlinks,allcolors=cvprblue]{hyperref}
\usepackage{siunitx}
\usepackage{graphicx}
\usepackage{subcaption}
\usepackage[export]{adjustbox}
\usepackage{cuted}
\usepackage{float}
\usepackage{pgfplots}
\usepgfplotslibrary{groupplots}
\pgfplotsset{compat=1.18}
  
\def\confName{CVPR}
\def\confYear{2027}

\title{End-to-End Image-Goal Navigation \\ for Agile Quadrotors via Recurrent State Distillation}

\author{Simone Nascivera \quad Ismail Geles \quad Davide Scaramuzza\\
Robotics and Perception Group, University of Zurich, Switzerland
}

\begin{document}
\setlength{\textfloatsep}{12pt plus 2pt minus 2pt}
\setlength{\dbltextfloatsep}{12pt plus 2pt minus 2pt}
\setlength{\intextsep}{8pt plus 2pt minus 2pt}
\maketitle

\begin{abstract}
Image-goal navigation asks a robot to reach the viewpoint of a target photograph without a map or goal pose.
In three dimensions, this requires not only recognizing the goal but also estimating 3D displacement and generating continuous control actions to approach and settle at the target view.
This is difficult because the same visual difference between current and goal image can require very different actions depending on the robot's velocity, yet monocular images alone reveal neither metric scale nor body motion.
As a result, existing image-goal policies are limited to planar action spaces and rely on a state estimator for flight control.
Policies that map directly from images to low-level control commands have been demonstrated for agile flight, but not toward goal images in unseen environments.
We present the first image-goal navigation policy that directly commands thrust and angular rates for agile quadrotor flight in unseen 3D environments, without relying on state estimation (e.g., SLAM, VO, or SfM) or a separate velocity controller.
We train a recurrent vision-based policy through a four-stage pipeline that combines geometric and motion supervision with privileged-teacher distillation and end-to-end reinforcement learning fine-tuning.
On 1000 held-out flights in unseen environments, the policy achieves an 82.3\% success rate with a mean terminal position error of 0.10\,m across successful flights.
Deployed zero-shot on a real quadrotor, the policy reaches an 84\% success rate over 25 trials in an unseen indoor environment.
\end{abstract}

\begin{figure}[t]
  \centering
  \includegraphics[width=\linewidth]{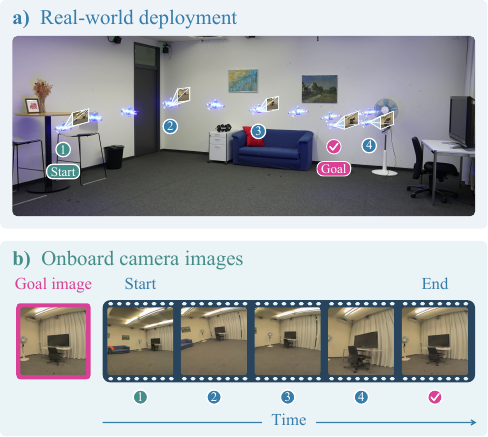}
  \caption{\textbf{A goal image specifies where to fly.}
  Real-world deployment of our method (a). 
  The quadrotor flies from the starting image to the viewpoint goal, only relying on the onboard camera frames (b). No explicit state estimation (i.e., SLAM, VO, or SfM) is used.}
  \label{fig:task}
  \vspace{-4pt}
\end{figure}

\begin{figure*}[t]
  \centering
  \includegraphics[]{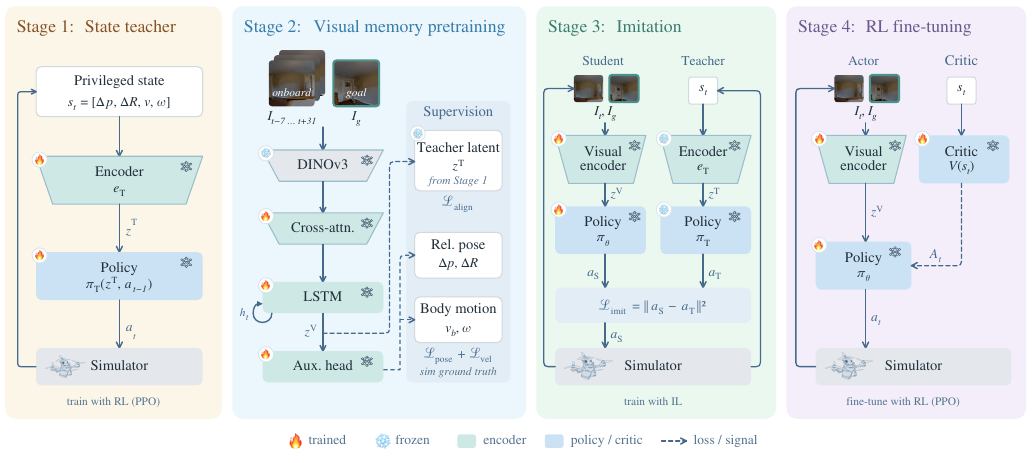}
  \caption{\textbf{Four-stage training pipeline.} The policy is trained progressively through state-based teacher learning (Stage 1), visual representation pretraining (Stage 2), imitation learning (Stage 3), and reinforcement learning fine-tuning (Stage 4). Stage 3 weights are initialized with the pretrained visual encoder and the teacher actor's weights, while Stage 4 fine-tunes the resulting policy for improved flight performance.
  }\label{fig:method}
\vspace{-10pt}
\end{figure*}

\section{Introduction}
\label{sec:intro}

Human pilots navigate quadrotors through unfamiliar environments using only an onboard camera stream, without explicitly estimating the vehicle's position, velocity, or attitude~\cite{geles2024pixels}.
We investigate whether autonomous flight can similarly rely on visual feedback alone: given a goal photograph, a quadrotor must reach the viewpoint from which it was taken as quickly as possible (\cref{fig:task}).
The photograph specifies a destination without requiring a map or a metric goal pose.
We study how to achieve this with a policy that maps visual observations and last applied actions directly to low-level control commands, without explicit state estimation such as Simultaneous Localization and Mapping (SLAM), Visual Odometry (VO), or Structure from Motion (SfM)~\cite{slam-handbook}.

Extracting the information needed for flight control from onboard frames and policy's past actions is challenging.
Reaching a goal viewpoint requires translating in three dimensions, aligning the camera, and braking to hover at the destination.
The appropriate control command depends not only on the displacement to the goal but also on the vehicle's current motion.
However, velocity cannot be directly inferred from a single frame, and monocular observations leave metric scale ambiguous.

To address these limitations, we use a recurrent policy that integrates visual information over time, allowing it to implicitly infer the vehicle's motion \cite{wang2019recurrent}.
The previous control command provides additional information about the vehicle dynamics, helping reduce metric scale ambiguity~\cite{ludhiyani2019multirotor}.
Tilting the vehicle also changes its thrust direction and camera view together.
A visual policy must account for these coupled changes over time.

Recent work, such as SIGN~\cite{yan2025sign}, has explored aerial image-goal navigation.
While the authors demonstrate image-goal navigation on quadrotors, including real-world flight, the method outputs planar velocity commands, relying on a separate state-estimation-based controller for flight stabilization.
In parallel, vision-based policies have demonstrated agile flight by directly commanding low-level control inputs, without explicit state estimation~\cite{geles2024pixels,heeg2025learning}.
Combining image-goal navigation with such direct flight control remains an open challenge, particularly when requiring three-dimensional motion, precise settling at the goal, and generalization to unseen real-world environments.

We address this problem through a four-stage training pipeline (\cref{fig:method}).
First, a state-based teacher learns to fly using privileged goal-relative pose and motion information.
We then train a visual representation from sequences of onboard images and a goal photograph, using geometric and motion supervision together with alignment to the teacher's latent representation.
During the next stage, the student is trained through imitation learning using the teacher from the first stage.
Finally, reinforcement learning fine-tunes the student policy in closed loop.
At deployment, the policy receives only the current image, the goal photograph, and its previous command, with no access to position, velocity, attitude, or IMU measurements.

\paragraph{Contributions.}
\textbf{(1)} To our knowledge, we present the first image-goal flight policy that directly commands low-level control inputs in unfamiliar environments without explicit state estimation.
\textbf{(2)} We introduce a four-stage training procedure that combines geometric and motion supervision, latent representation alignment, imitation learning, and reinforcement learning to learn a vision-based flight policy.
\textbf{(3)} We evaluate the policy on 1000 held-out flights in 100 unseen environments, achieving an 82.3\% success rate with a mean terminal position error of 0.10\,\unit{m} across successful flights, and demonstrate zero-shot transfer to a real quadrotor with an 84\% success rate over 25 trials in an unseen indoor environment.

\section{Related Work}
\label{sec:related}

\paragraph{Aerial image-goal navigation.}
SIGN~\cite{yan2025sign} and Omni-IGN~\cite{yan2026omniign}
demonstrate aerial image-goal navigation with planar velocity
commands and dedicated flight-control loops.
Their deployed systems use visual-inertial and LiDAR-inertial
odometry, respectively, for low-level control.
Both address goal discovery and collision avoidance, with
Omni-IGN extending perception to four fisheye views.
UA-NWM~\cite{zhu2026uanwm} uses a latent world model to score
candidate trajectories and plan waypoints, while
DreamNav~\cite{li2026dreamnav} refines relative-motion predictions
by generating candidate observations.
Our setting instead focuses on reaching and settling at covisible
goal viewpoints under quadrotor dynamics, including altitude
changes.
The policy directly commands collective thrust and body rates
from visual observations and previous commands, without explicit
vehicle-state inputs or a separate velocity controller.

\paragraph{Image-goal representations.}
Image-goal navigation has been extensively studied in simulated
embodied environments~\cite{zhu2017target,yadav2023ovrlv2,li2025regnav}.
FGPrompt~\cite{sun2023fgprompt} emphasizes fine-grained goal
conditioning, while Monaci~\etal~\cite{monaci2025matter} analyze
visual fusion strategies and relate navigation performance to
the relative geometry recoverable from learned representations.
DEBiT~\cite{bono2024debit} pretrains a binocular transformer
through cross-view completion and relative-pose and visibility
supervision, then integrates its features into a recurrent
navigation policy.
We similarly use current--goal cross-attention, but pretrain
the recurrent representation on ordered flight sequences with
geometric and motion supervision.
Complementary approaches such as
AnyImageNav~\cite{deng2026anyimagenav} combine exploration with
explicit geometric registration of the goal image using depth
and pose observations.
Our deployed policy instead uses the learned representation
directly for flight control.

\paragraph{Visual servoing and relative-pose estimation.}
Classical image-based visual servoing
(IBVS)~\cite{chaumette2006visual} computes camera-velocity
commands from image-feature errors.
Learned representations can improve feature matching, as in
ViT-VS~\cite{scherl2025vitvs}, which combines pretrained
DINOv2 correspondences with an IBVS control law.
However, avoiding explicit pose reconstruction in the visual
front end does not necessarily eliminate state estimation
from the complete flight-control pipeline.
In the velocity-command IBVS baselines considered here,
a separate controller uses vehicle-state estimates to track
the commanded motion.
Learned matching and relative-pose methods, including
MASt3R~\cite{leroy2024mast3r} and
Reloc3r~\cite{dong2025reloc3r}, provide geometric estimates
that can likewise support modular navigation systems.

Our objective is instead to perform image-goal flight without
explicit vehicle-state estimation or an intermediate
velocity-control loop.
The recurrent policy maps visual observations and previous
commands directly to collective thrust and body rates,
learning to account for the coupling between vehicle attitude,
translation, and camera motion.

\paragraph{Privileged learning for visual control.}
Teacher--student learning transfers privileged training
information to deployable visuomotor
policies~\cite{chen2019lbc,kaufmann2020acrobatics,loquercio2021wild}.
Xing~\etal~\cite{xing2024bootstrapping} combine state-based
teacher training, visual imitation learning, and reinforcement
learning fine-tuning for agile flight.
We adopt this progression and introduce a separate visual
representation-learning stage before action imitation.
Our latent alignment is related to
RMA~\cite{kumar2021rma}, which predicts privileged environment
and dynamics embeddings from observation--action histories,
and perceptive locomotion methods that learn recurrent
representations with privileged
supervision~\cite{miki2022learning}.
Here, the visual representation is supervised by goal-relative
pose and body motion and aligned with the teacher's latent
space to support reuse of its actor.

\paragraph{Vision-based low-level flight control.}
Geles~\etal~\cite{geles2024pixels} demonstrate agile flight
without explicit state estimation using gate-based visual
representations, while
Dream to Fly~\cite{romero2026dream} learns flight control
from pixels through model-based reinforcement learning.
Heeg~\etal~\cite{heeg2025learning} study differentiable
training of quadrotor controllers from visual features.
Gaussian-splatting environments also support visuomotor
policy learning through differentiable
dynamics~\cite{chen2025gradnav} or imitation of an MPC
expert~\cite{low2024sousvide}, with
GRaD-Nav++~\cite{chen2025gradnavpp} introducing language
conditioning.
Our work addresses image-goal-conditioned flight: the target
viewpoint is specified by a photograph, and the policy must
reach and settle at that viewpoint in unseen environments.
Training does not require gradients through the simulator.

\begingroup
\setlength{\abovedisplayskip}{8pt plus 2pt minus 4pt}
\setlength{\belowdisplayskip}{8pt plus 2pt minus 4pt}
\section{Method}
\label{sec:method}

\Cref{fig:method} illustrates the four stages of our training pipeline and their main components.

\subsection{Task Definition}
The objective is to fly a quadrotor quickly and precisely from an initial stationary pose to a goal viewpoint.
The goal is specified by an RGB image $I_g$, captured at a viewpoint with partial visual overlap with the initial observation and zero roll and pitch.
At each time step $t$, the policy receives the current image $I_t$, the goal image $I_g$, and the previous normalized control command $a_{t-1}$.
The policy is recurrent, with hidden state $h_{t-1}$, and outputs
\begin{equation}
  a_t=\pi_\theta(I_t,I_g,a_{t-1};h_{t-1}),
  \qquad a_t\in[-1,1]^4.
\end{equation}
The normalized action is mapped to collective thrust per unit mass and three body angular rates.
Notably, the policy receives no explicit state estimates such as position, velocity, or attitude.
A flight is considered successful when the quadrotor reaches the goal
viewpoint within specified position, orientation, and velocity
tolerances defined in section~\cref{sec:experiments}.

\subsection{Quadrotor Simulator}

We simulate the quadrotor dynamics using the standalone Warp-based simulator described in~\cite{zhai2026learning}, derived from the Agilicious architecture~\cite{foehn2022agilicious}.
Our custom quadrotor has a mass of $0.21\,\mathrm{kg}$, an arm length of $19.4\,\mathrm{cm}$, a propeller radius of $3.81\,\mathrm{cm}$, and a thrust-to-weight ratio of $6.8$.
The simulator models calibrated inertia, motor thrust characteristics, battery dynamics, aerodynamic effects, and actuation delays, with a nominal command delay of $40\,\mathrm{ms}$.
The policy outputs collective thrust and body-rate (CTBR) commands, which are converted into motor commands by the Betaflight low-level controller.
The dynamics run at 500\,Hz, while the policy operates at 50\,Hz, holding each command constant for 10 simulation steps.

Visual observations are generated using Isaac Lab's camera interface and NVIDIA RTX renderer, with the rendered pose synchronized to the Warp dynamics.
A forward-facing monocular camera mounted at the body-frame origin produces $224\times224$ images using a calibrated polynomial fisheye model.
Depth is additionally rendered for training-data validation and reward computation, but is never provided to the policy.

\subsection{Stage 1: State-Based Teacher}

We first train a state-based teacher using PPO~\cite{schulman2017ppo}.
Let $\mathbf{p}_t$, $R_t$, $\mathbf{v}_t$, and $\boldsymbol{\omega}_t$ denote the position, body-to-world orientation, world-frame linear velocity, and body angular velocity, respectively.
The teacher observes the 12-dimensional goal-relative state
\begin{equation}
\begin{aligned}
s_t = \big[&
R_t^\top(\mathbf{p}_g-\mathbf{p}_t),\;
\log(R_t^\top R_g)^\vee,\\
& R_t^\top\mathbf{v}_t,\;
\boldsymbol{\omega}_t
\big]\in\mathbb{R}^{12}.
\end{aligned}
\label{eq:state}
\end{equation}
Here, $\log$ denotes the matrix logarithm on $\mathrm{SO}(3)$, and $(\cdot)^\vee$ maps a skew-symmetric matrix to its corresponding 3D rotation vector.
A linear layer followed by GELU embeds $s_t$ into a 128-dimensional representation $z_t^T=e_T(s_t)$.
The actor concatenates $z_t^T$ with the previous action $a_{t-1}$ and passes them through two 512-unit hidden layers to output four normalized commands in $[-1,1]$.
The privileged critic additionally observes the flattened rotation matrix, yielding a 25-dimensional input $[s_t,a_{t-1},\operatorname{vec}(R_t)]$.

The reward combines position progress, attitude alignment, height tracking, control regularization, and terminal outcomes:
\begin{equation}
\begin{aligned}
r_t ={}& r_t^{\mathrm{pos}}+r_t^{\mathrm{att}}
+r_t^{\mathrm{height}}+r_t^{\mathrm{reg}}\\
&-0.005+10\mathbb{I}_t^{\mathrm{success}}
-20\mathbb{I}_t^{\mathrm{failure}}.
\end{aligned}
\end{equation}
Defining $d_t=\|\mathbf{p}_t-\mathbf{p}_g\|_2$ and $\Delta d_t=d_{t-1}-d_t$, the task rewards are
\begin{equation}
\begin{aligned}
r_t^{\mathrm{pos}} &=
\begin{cases}
4\Delta d_t, & \Delta d_t\geq 0,\\
20\Delta d_t, & \Delta d_t<0,
\end{cases}\\
r_t^{\mathrm{att}} &= 0.125(\theta_{t-1}-\theta_t),\\
r_t^{\mathrm{height}} &=
-0.1(p_{t,z}-z_t^{\mathrm{ref}})^2.
\end{aligned}
\end{equation}
Here, $\theta_t$ is the geodesic orientation error relative to a look-at frame pointing 2\,m beyond the goal along its forward axis, with roll determined by the goal's up direction.
The reference height $z_t^{\mathrm{ref}}$ follows the linear interpolation between the start and goal heights according to horizontal progress.
For predominantly vertical flights, it instead penalizes motion beyond the endpoint heights.

Control regularization penalizes changes in thrust and body-rate commands, as well as excessive angular rates:
\begin{equation}
\begin{aligned}
r_t^{\mathrm{reg}}=-\lambda_\ell\big[&
0.3(\Delta\bar{u}_{T,t})^2\\
&+0.05\|\Delta\bar{\mathbf{u}}_{\omega,t}\|_2^2\\
&+0.05\|\bar{\boldsymbol{\omega}}_{xy,t}\|_2^2\\
&+0.025\|\bar{\mathbf{u}}_{\omega,t}\|_2^2
\big].
\end{aligned}
\end{equation}
Here, $\Delta\bar{\mathbf{u}}_t$ denotes command differences normalized by their respective scales, while $\bar{\boldsymbol{\omega}}_t$ and $\bar{\mathbf{u}}_{\omega,t}$ are normalized by the body-rate limits.
The curriculum multiplier $\lambda_\ell$ increases from $0.25$ to $1.25$.

We adopt a 20 discrete-steps curriculum that progressively increases goal displacement, yaw differences, and dynamics perturbations.
Advancement requires at least $90\%$ success over an assessment window of 1024 completed episodes, aggregated across parallel environments.
Success requires position, velocity, and attitude errors to remain within stage-specific tolerances for a prescribed number of consecutive control steps.
We additionally randomize mass, inertia, drag, actuation parameters, and delays.
The resulting teacher learns goal-directed flight control from privileged state observations, providing the target behavior and latent representation for subsequent stages.

\subsection{Stage 2: Visual Encoder}

In the second stage, we train a visual encoder to infer the 12-dimensional privileged pose-and-motion state $s_t$, defined in \cref{eq:state}, from the current image $I_t$, goal image $I_g$, and previous action $a_{t-1}$.
A frozen DINOv3 ViT-S/16 backbone~\cite{simeoni2025dinov3} extracts 384-dimensional patch descriptors from both images.
Similar to~\cite{jang2025pow3r}, we augment each patch with its corresponding 3D camera ray.
The visual descriptors are projected to 256 dimensions and combined by elementwise addition with ray embeddings produced by a learned $3\rightarrow64\rightarrow256$ MLP:
\begin{equation}
\tilde{\mathbf f}_i =
\operatorname{Adapter}(\mathbf f_i)
+\operatorname{MLP}_{\mathrm{ray}}(\mathbf r_i).
\end{equation}
The resulting 256-dimensional tokens are processed by two bidirectional cross-attention blocks, allowing current and goal patches to exchange information.
Learned spatial pooling and fusion of the two streams, their difference, and elementwise product produce a 128-dimensional pair feature $x_t$.
The previous normalized command $a_{t-1}\in[-1,1]^4$ is concatenated with this feature before being passed to a single-layer LSTM:
\begin{equation}
\begin{aligned}
\tilde{x}_t &= [x_t;a_{t-1}],\\
(h_t,c_t) &= \operatorname{LSTM}(\tilde{x}_t,h_{t-1},c_{t-1}),\\
z_t^V &= P(h_t)\in\mathbb{R}^{128},
\end{aligned}
\end{equation}
where $P$ is a linear projection followed by GELU.
The LSTM has a hidden dimension of 256, while the projected representation $z_t^V$ has dimension 128, matching the teacher's latent space.
The recurrent state is reset at episode boundaries and goal changes.
An auxiliary head predicts the privileged state $\hat{s}_t=H(z_t^V)$.

We train the encoder using geometric and motion supervision, together with alignment to the frozen teacher representation:
\begin{equation}
\begin{aligned}
\mathcal L_{\mathrm{pre}} ={}&
\lambda_p\Big(
\|\hat p_{\mathrm{rel}}-p_{\mathrm{rel}}\|_2^2\\
&+d_{\mathrm{SO}(3)}(\hat R_{\mathrm{rel}},R_{\mathrm{rel}})^2
\Big)\\
&+\frac{\lambda_v}{3}\|\hat v_b-v_b\|_2^2\\
&+\frac{\lambda_\omega}{3}\|\hat\omega-\omega\|_2^2\\
&+\frac{\lambda_z}{128}
\|z_t^V-\operatorname{sg}(z_t^T)\|_2^2.
\end{aligned}
\label{eq:pretraining}
\end{equation}
Here, $z_t^T=e_T(s_t)$ is the frozen teacher embedding, $\operatorname{sg}$ stops gradients, and $d_{\mathrm{SO}(3)}$ denotes the geodesic rotation error.
The predicted rotation vector is converted into $\hat R_{\mathrm{rel}}$ through the exponential map before computing the rotation loss.
Position, orientation, linear velocity, and angular velocity errors use reference scales of 1\,m, 1\,rad, 1\,m/s, and 1\,rad/s, respectively.
All loss weights are set to one.
The teacher encoder and DINOv3 backbone remain frozen.

Training uses ordered clips from teacher rollouts, with eight warm-up steps to initialize the recurrent state without gradients, followed by 32 supervised steps.
We collect clips from teacher curriculum levels $\{0,4,8,19\}$ and weight their contributions equally in the training objective.
Losses are averaged over valid supervised frames after warm-up.
The resulting representation is trained to match the teacher's latent features, allowing the pretrained actor to be reused in the subsequent imitation learning stage.

\subsection{Stage 3: Imitation Learning}

In the third stage, we initialize the student actor with the teacher's weights and replace the teacher's state-based embedding $z_t^T$ with the pretrained recurrent visual embedding $z_t^V$.
The actor receives $z_t^V$ and the previous normalized command $a_{t-1}\in[-1,1]^4$, without explicit state estimates.

Following the dataset-aggregation principle of DAgger~\cite{ross2011dagger}, the teacher provides action labels at states visited during data collection.
At each control step, we execute the teacher's action with probability $\beta_k$ and the student's action otherwise.
The teacher-selection probability decreases linearly from $0.5$ to zero over 1,000 training iterations.
Teacher supervision continues after the student assumes full control.

We use the same 20-level competence-based curriculum as in Stage 1.
Curriculum advancement is suspended during teacher-assisted collection.
Once assistance ends, the curriculum statistics are reset, and advancement requires a student-only success rate of at least $90\%$ over at least 1024 completed episodes.
Collected trajectories are stored as ordered replay clips, preserving temporal order and episode boundaries.
Each clip contains 8 burn-in steps, processed without gradients to reconstruct recurrent memory, followed by 32 supervised steps.
The student minimizes
\begin{equation}
    \mathcal L_{\mathrm{imit}} =
    \mathbb E_{\mathcal D}\left[
        \frac{1}{4}
        \left\|
        \pi_\theta(z_t^V,a_{t-1})-a_t^T
        \right\|_2^2
        \right],
\end{equation}
where $a_t^T$ is the teacher's action at the visited state.
The loss is averaged over valid supervised frames after burn-in.
Gradients jointly update the visual attention encoder, LSTM, projection, and actor, while the DINOv3 backbone remains frozen.
No auxiliary state-prediction or embedding-alignment losses are applied during this stage.
We disable collision detection during imitation learning, as we empirically found it to hinder training, and re-enable it during the subsequent reinforcement learning stage.

\subsection{Stage 4: Reinforcement Learning}

In the final stage, we initialize the actor and recurrent visual encoder from Stage 3, while the privileged critic retains the Stage 1 teacher's weights, unchanged during imitation learning.
We jointly fine-tune the visual attention encoder, LSTM, projection, and actor using PPO, while keeping DINOv3 frozen.
The critic is optimized using privileged simulator state.
All trainable components, including the critic, share a fixed learning rate of $10^{-5}$ through a single Adam optimizer, without a separate critic warm-up.

We retain the Stage 1 reward and introduce an additional image-reprojection penalty near the goal:
\begin{equation}
r_t^{(4)} = r_t^{(1)} + r_t^{\mathrm{repr}}.
\end{equation}
The penalty measures image alignment by projecting 3D points from the goal image into the current view using ground-truth simulator geometry.
It combines reprojection error and geometric image overlap, activates within $0.5\,\mathrm{m}$ of the goal, and is bounded by $0.02$ per control step.
The penalty smoothly vanishes at the activation boundary.
The simulator geometry used to compute this penalty is required only during training.
We also re-enable collision detection, which was disabled during imitation learning.
The complete reward formulation is provided in the supplementary material.

\subsection{Training Details}

We precompute approximately 10,000 start-goal pairs per HM3D training building, requiring zero tilt, relative yaw $\leq90^\circ$, covisibility $\geq20\%$, valid depth coverage $\geq70\%$, and 0.6\,m clearance along the straight segment, including both endpoints.
The Stage 1 curriculum separately allows yaw differences up to $180^\circ$.
Valid pairs are cached to avoid repeated geometric checks during training.

All stages are trained on a single NVIDIA RTX 5090 GPU for 10,000, 10,000, 25,000, and 35,000 iterations, respectively.
Stage 1 uses 2,048 parallel environments with 24 rollout steps (49,152 transitions per iteration), five PPO epochs, four mini-batches, and an adaptive learning rate initialized at $10^{-3}$ (target KL divergence of 0.02).
Stages 2 to 4 use 256 parallel environments.
Stages 2 and 3 optimize mini-batches of 128 and 1,024 supervised frames, respectively, using 32-step clips with eight additional burn-in steps and masking invalid frames.
Both use a fixed learning rate of $10^{-4}$.
Stage 4 collects 6,144 transitions per iteration, with three PPO epochs, 16 mini-batches, and a fixed learning rate of $10^{-5}$.

\endgroup
\section{Experiments}
\label{sec:experiments}

\begin{table*}[!htbp]
    \centering
    \caption{
HM3D validation results on 1,000 trajectories across 100 held-out buildings.
Our method and IBVS use a 20\,s time limit, while SIGN uses 500\,s to account for its lower maximum velocity.
Completion time and final errors are reported only for successful flights.
For SIGN, we report two success rates, with the value in parentheses computed using the success criterion from the original paper~\cite{yan2025sign}.
Bold indicates the best results among non-oracle methods.}
    \label{tab:hm3d}
    \footnotesize
    \setlength{\tabcolsep}{4pt}
    \begin{tabular*}{\linewidth}{@{\extracolsep{\fill}}lcccccc@{}}
        \toprule
        \textbf{Method}
        & \textbf{Velocity [m/s]}
        & \textbf{Success [\%]} $\uparrow$
        & \textbf{Failures} $\downarrow$
        & \textbf{Time [s]} $\downarrow$
        & \textbf{Position error [m]} $\downarrow$
        & \textbf{Attitude error [$^\circ$]} $\downarrow$ \\
        & \textbf{Limit}
        & \textbf{Rate}
        & \textbf{Crashes / Timeouts}
        & \textbf{Mean / Median}
        & \textbf{Mean / Median}
        & \textbf{Mean / Median} \\
        \midrule
        Oracle IBVS & 1 & 74.6 & 252 / 2
        & 3.7 / 3.3 & 0.04 / 0.04 & 3.31 / 2.03 \\
        & 3 & 70.1 & 299 / 0
        & 2.4 / 2.3 & 0.04 / 0.03 & 2.76 / 2.71 \\
        & 5 & 66.7 & 333 / 0
        & 2.3 / 2.3 & 0.04 / 0.03 & 2.77 / 2.45 \\
        \midrule
        SIFT IBVS & 1 & 40.8 & \textbf{26} / 566
        & 2.9 / 2.5 & \textbf{0.04} / 0.04 & 3.43 / 3.52 \\
        & 3 & 38.8 & 52 / 560
        & \textbf{2.4} / \textbf{2.1} & \textbf{0.04} / \textbf{0.03} & \textbf{2.56} / 2.29 \\
        & 5 & 38.2 & 61 / 557
        & 2.5 / \textbf{2.1} & \textbf{0.04} / \textbf{0.03} & 2.64 / \textbf{2.28} \\
        SIGN & 0.25 & 1.1 / (53.0) & 71 / 918
        & 66.5 / 25.9 & 0.12 / 0.14 & 7.96 / 9.74 \\
        Ours & -- & \textbf{82.3} & 89 / \textbf{88}
        & 5.1 / 4.7 & 0.10 / 0.11 & 2.77 / 2.52 \\
        \bottomrule
    \end{tabular*}
    \vspace{-6pt}
\end{table*}

\begin{table*}[!htbp]
    \caption{Gibson evaluation on 1,000 matched validation flights, with a time limit of 500\,s for all methods.
        Time and error statistics summarize successful flights only.
        For SIGN, we report two success rates, with the value in parentheses computed using the success criterion from the original paper~\cite{yan2025sign}.
        Bold indicates the best results among non-oracle methods.
        }
    \label{tab:gibson_hold1}
    \centering
    \footnotesize
    \setlength{\tabcolsep}{4pt}
    \begin{tabular*}{\linewidth}{@{\extracolsep{\fill}}lcccccc@{}}
        \toprule
        \textbf{Method}
        & \textbf{Velocity [m/s]}
        & \textbf{Success [\%]} $\uparrow$
        & \textbf{Failures} $\downarrow$
        & \textbf{Time [s]} $\downarrow$
        & \textbf{Position error [m]} $\downarrow$
        & \textbf{Attitude error [$^\circ$]} $\downarrow$ \\
        & \textbf{Limit}
        & \textbf{Rate}
        & \textbf{Crashes / Timeouts}
        & \textbf{Mean / Median}
        & \textbf{Mean / Median}
        & \textbf{Mean / Median} \\
        \midrule
        Oracle IBVS & 1 & 95.7 & 42 / 1
        & 2.6 / 2.3 & 0.03 / 0.02 & 3.18 / 2.13 \\
        & 3 & 92.0 & 79 / 1
        & 2.1 / 2.0 & 0.03 / 0.03 & 2.94 / 2.67 \\
        & 5 & 91.5 & 84 / 1
        & 2.1 / 2.0 & 0.03 / 0.03 & 2.95 / 2.61 \\
        \midrule
        SIFT IBVS & 1 & 80.1 & 67 / 132
        & 6.4 / 2.2 & \textbf{0.03} / \textbf{0.02} & 3.27 / 2.48 \\
        & 3 & 74.5 & 136 / 119
        & \textbf{3.4} / \textbf{2.0} & \textbf{0.03} / 0.03 & 2.89 / 2.69 \\
        & 5 & 74.6 & 136 / 118
        & \textbf{3.4} / \textbf{2.0} & \textbf{0.03} / 0.03 & 2.94 / 2.65 \\
        SIGN & 0.25 & 11.7 / (79.0) & \textbf{0} / 883
        & 24.3 / 23.2 & 0.10 / 0.11 & 7.89 / 9.80 \\
        Ours & -- & \textbf{93.0} & 53 / \textbf{17}
        & 5.4 / 4.0 & 0.10 / 0.10 & \textbf{2.53} / \textbf{2.35} \\
        \bottomrule
    \end{tabular*}
    \vspace{-6pt}
\end{table*}

We evaluate our policy's ability to reach goal viewpoints in buildings unseen during training.
We compare against classical and learning-based methods in simulated environments and evaluate zero-shot transfer to a real-world setting.
Finally, we conduct ablation studies to assess the contribution of the main components of our approach.

\vspace{-4pt}
\subsection{Experimental Setup}
\label{sec:experimental_setup}
\paragraph{Datasets}
We train on the 800 HM3D training buildings~\cite{ramakrishnan2021hm3d} and evaluate on 1,000 fixed flights across the 100 validation buildings, with 10 flights per building.
The flights are divided into three goal-distance ranges: 334 at 0.3-2\,\unit{m}, 333 at 2-5\,\unit{m}, and 333 at 5-10\,\unit{m}.
Rather than using the trajectories provided by the dataset, we sample our own start-goal pairs to satisfy the covisibility requirements of our task.
Specifically, we require at least 20\% visual overlap between the initial and goal views, a maximum relative yaw of 90 degrees, and zero roll and pitch at both poses.
We require 0.6\,\unit{m} clearance along the straight start-goal segment, endpoints included. This pair filter does not constrain the executed flight path.
All validation buildings are unseen during training.
To further assess generalization, we evaluate the policy zero-shot on 1,000 fixed flights across 14 Gibson~\cite{xia2018gibson} buildings, none of which are used for training.

Finally, we evaluate the policy zero-shot on a physical quadrotor using its onboard camera in an indoor flight arena.
We use 25 start-goal image pairs to assess real-world flight performance.

\paragraph{Baselines.}
To our knowledge, no existing image-goal navigation method directly commands collective thrust and body rates without explicit state estimation. 
We therefore compare against methods that rely on state estimates for low-level flight control, despite the additional information available to them.

We compare against two variants of Image-Based Visual Servoing (IBVS)~\cite{chaumette2006visual}.
\emph{Oracle IBVS} uses ground-truth correspondences between the current and goal images, while \emph{SIFT IBVS} obtains correspondences using SIFT features~\cite{lowe2004distinctive} and brute-force descriptor matching.
Both variants use ground-truth depth for the current image and noise-free state estimates, with Oracle IBVS additionally using the goal-image depth.
We evaluate both variants with translational velocity command limits of 1, 3, and 5\,\unit{m/s}, noting that the achieved flight speed may exceed these limits.
The resulting camera velocity and yaw-rate commands are passed to a geometric velocity and attitude controller, which converts them into collective thrust and body-rate commands.
Both baselines use the same quadrotor, camera configuration, and low-level controller as our method.

We also evaluate SIGN~\cite{yan2025sign} using its native camera configuration and a 1\,\unit{s} decision interval. 
SIGN outputs forward velocity and yaw-rate commands, without vertical control.
As with the IBVS baselines, we use a geometric controller with access to the ground-truth simulator state to convert these commands into collective thrust and body rates, ensuring the same low-level control interface and as our method.

\paragraph{Evaluation Metrics.}
A flight succeeds when position error is below 0.15\,m, attitude error below 10\,$^\circ$, and speed below 0.08\,\unit{m/s} for one control step (0.02\,\unit{s}).
Flights that neither succeed nor crash within the time limit are counted as timeouts.
To detect collisions, we sweep a body-centered sphere of radius 0.2\,\unit{m} along the executed trajectory, sampling at intervals of at most 0.05\,\unit{m}, and count contact with the building mesh as a crash.
We report the success, crash, and timeout rates across the evaluation set. 
For successful flights, we additionally report the mean completion time [\unit{s}], final attitude error [$^\circ$], and final position error [\unit{m}].

\vspace{-4pt}
\subsection{Synthetic Evaluation}
\paragraph{HM3D.}
\label{par:results_hm3d}

Table~\ref{tab:hm3d} reports the performance of our method and the baselines on the HM3D validation set.
Our method achieves the highest success rate among all baselines, despite operating without explicit state estimation or a velocity controller.
However, it exhibits longer completion times and larger final position errors than the IBVS baselines.
Oracle IBVS achieves a significantly higher success rate than SIFT IBVS, but also suffers from more collisions.
In contrast, SIFT IBVS predominantly fails through timeouts.
Over 99\% of these timed-out flights have a terminal speed below 0.003\,\unit{m/s}; their median remaining distance to the goal is 5\,\unit{m}.
This suggests that unreliable feature correspondences cause the controller to stall before reaching the goal.
For SIGN, we report both the success rate under our criteria and the one presented in the paper itself, which is within 1\,\unit{m} and 25\,$^\circ$ from the target pose. 
In addition, in this secondary success rate we consider a 2D projected distance, instead of 3D like all the other baselines, since the original paper maintained constant altitude.

\paragraph{Gibson.}
\label{par:results_gibson}

Our policy demonstrates strong generalization to the unseen Gibson dataset, achieving a success rate of 93\%.
As on HM3D, Table~\ref{tab:gibson_hold1} shows that collisions are the dominant failure mode for the Oracle IBVS baseline, becoming more frequent as the velocity command limit increases.
SIFT IBVS instead fails due to both collisions and timeouts.
SIGN achieves a higher success rate than on HM3D.
However, the trajectories sampled for both synthetic evaluations do not enforce the constant-altitude constraint used during SIGN's training.
Since SIGN does not command vertical motion, we include it as a reference for existing image-goal navigation methods rather than a directly comparable solution to our three-dimensional navigation task.
Finally, our method remains the only evaluated policy that does not rely on explicit state estimation or a separate velocity controller.

\vspace{-4pt}
\subsection{Real World Deployment}
We deploy our policy zero-shot in an indoor flight arena, following a setup similar to \cite{geles2024pixels,zhai2026learning}.
We use a modified version of \textit{Agilicious}~\cite{foehn2022agilicious}, streaming control commands from a workstation to the quadrotor via an RF transmitter and receiving low-latency video from its onboard camera.
The policy is optimized using NVIDIA TensorRT~\cite{nvidia_tensorrt} with FP32 precision, achieving a median inference time of 1.06\,\unit{ms} on an NVIDIA RTX 4090 GPU.

\begin{figure}[!htbp]
    \centering
    \begin{tabular}{@{}c@{\hspace{2pt}}ccc@{}}
        \adjustbox{valign=m}{\rotatebox{90}{Goal Viewpoint}} &
        \includegraphics[width=0.28\columnwidth,valign=m]{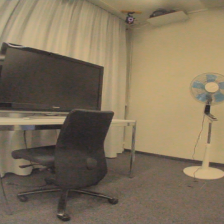} &
        \includegraphics[width=0.28\columnwidth,valign=m]{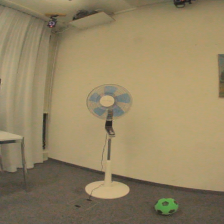} &
        \includegraphics[width=0.28\columnwidth,valign=m]{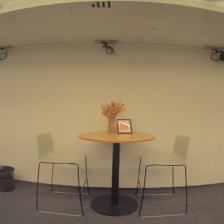} \\

        \rule{0pt}{5pt} & & & \\
        \adjustbox{valign=m}{\rotatebox{90}{Start Viewpoint}} &
        \includegraphics[width=0.28\columnwidth,valign=m]{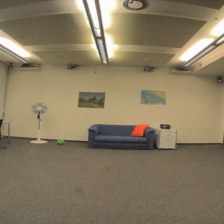} &
        \includegraphics[width=0.28\columnwidth,valign=m]{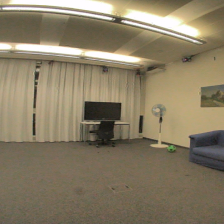} &
        \includegraphics[width=0.28\columnwidth,valign=m]{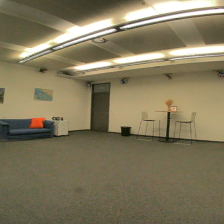} \\[-10pt]
        
        & \subcaptionbox{Sequence 0}[0.28\columnwidth]{\strut}
        & \subcaptionbox{Sequence 10}[0.28\columnwidth]{\strut}
        & \subcaptionbox{Sequence 11}[0.28\columnwidth]{\strut}
    \end{tabular}

    \caption{Three examples of the 25 start and goal viewpoints used during the real world experiments in the indoor flying arena.}
    \label{fig:real_world_start_goal}
    \vspace{-6pt}
\end{figure}

Figure~\ref{fig:real_world_start_goal} shows a subset of the 25 start-goal image pairs used in our real-world experiments.
We apply the same success criteria as in simulation, while using a motion capture system only for recording the resulting performance.

Our policy achieves a success rate of \textbf{84\%}, with a mean final position error of \textbf{0.118\,\unit{m}} and orientation error of \textbf{3.07\,$^\circ$} across successful flights.
The initial position and orientation differences average 3.02\,\unit{m} and 18.95\,$^\circ$, respectively.
These results demonstrate that the policy transfers to a previously unseen real-world environment without fine-tuning, despite variable video transmission delays and substantial frame drops. Across the 25 experimental recordings, 30.51\% of received frames were dropped. 

\vspace{-4pt}
\subsection{Ablations}
\label{sec:ablations}
We evaluate the contribution of the recurrent visual encoder and the importance of each training stage.
For the visual encoder, we study the effect of previous control commands, latent dimension, and recurrence on geometric and motion prediction.
For the training pipeline, we compare different initialization strategies and assess the contribution of each training stage.

\begin{figure}[!htbp]
  \vspace{2pt}
  \centering
  \includegraphics[trim=-5pt 0pt 0pt 0pt,width=\columnwidth]{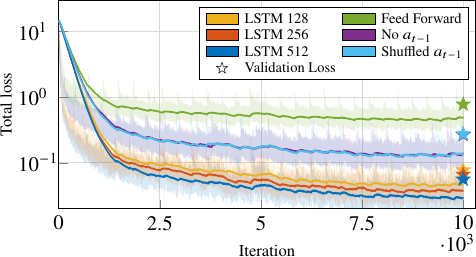}
  \caption{
  Stage 2 visual encoder ablations showing the evolution of training loss, along with the final validation loss. 
}
  \vspace{-14pt}
  \label{fig:stage2_loss}
\end{figure}

\vspace{-4pt}
\paragraph{Recurrent Visual Encoder.}
We ablate previous control commands, recurrence, and LSTM hidden dimension.
\Cref{fig:stage2_loss} shows training losses smoothed with an exponential moving average (decay 0.995), along with the validation loss evaluated at iteration 10000.
Validation uses 400 held-out trajectories from curriculum levels 0, 4, 8, and 19 across 100 unseen HM3D buildings.

Removing previous commands increases training and validation losses by factors of 3.7 and 4, respectively.
Shuffling command sequences across trajectories, while preserving teacher control and supervision, produces losses comparable to removing commands entirely.
This suggests that the benefit comes from the temporal correspondence between actions and visual observations, rather than command values alone.

Increasing the LSTM hidden dimension from 128 to 256 and 512 reduces validation loss from 0.078 to 0.066 and 0.056, respectively.
We select 256 to limit computational cost, as 512 increases training time by 11\%.
Finally, removing recurrence increases validation loss by nearly $12\times$, with angular velocity prediction loss increasing by $23.6\times$, highlighting the importance of visual memory for motion inference.

\vspace{-8pt}
\paragraph{Staged Training.}
We evaluate the contribution of each training stage on the 1,000 held-out HM3D trajectories used in our main evaluation.
The student after imitation learning (Stage 3) achieves a success rate of 64.8\%, which increases to 82.3\% after reinforcement learning fine-tuning (Stage 4).

To assess the importance of initialization, we train PPO directly from scratch, starting at curriculum level 0 with an adaptive learning rate initialized at $10^{-4}$ and a target KL divergence of 0.01.
This variant fails to progress beyond the initial curriculum level.
Initializing PPO from the pretrained visual encoder (Stage 2), without imitation learning, allows limited progress but fails to advance beyond the first curriculum level.
Similarly, training Stage 3 without Stage 2 visual pretraining fails to progress beyond the initial curriculum level.

These results highlight the importance of staged initialization: visual pretraining enables imitation learning to progress through the curriculum, while imitation provides an effective initialization for reinforcement learning, which further improves flight success.
\begin{figure}[!htbp]
  \vspace{-1pt}
  \centering
  \includegraphics[]{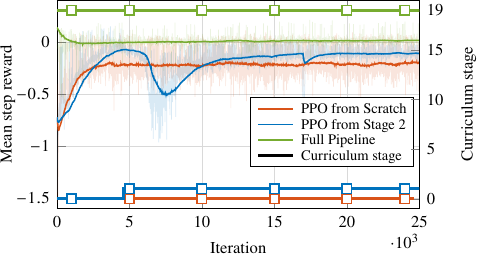}
  \caption{
Stage 4 initialization ablation. The full pipeline progresses through the curriculum, while scratch and Stage 2 initializations stall at early levels.
}
  \vspace{-10pt}
  \label{fig:stage4_loss}
\end{figure}

\vspace{-8pt}
\section{Conclusion}
We present the first image-goal navigation policy that enables quadrotors to reach a target viewpoint using onboard images and previous control commands, directly outputting collective thrust and body rates without explicit state estimation.
Our approach uses a four-stage training pipeline, progressing from a state-based teacher to a recurrent visual representation learned from foundation model features, followed by imitation learning and reinforcement learning fine-tuning.
Our ablations demonstrate the importance of staged training for efficient convergence and successful flight.
We evaluated the policy on held-out HM3D buildings and demonstrated zero-shot generalization to the Gibson dataset.
Finally, we deployed the policy on a physical quadrotor, achieving an 84\% success rate in a previously unseen real-world environment.

Our current setting assumes partial visual overlap between the initial and goal images and does not address exploration.
Future work will investigate how to combine goal-directed exploration with direct visual flight control to relax this assumption.

\section*{LLM Usage Statement}
OpenAI's ChatGPT and Anthropic's Claude were used to assist with language editing, improving manuscript clarity, and identifying relevant references. OpenAI Codex was used to refine figures and assist with code development, debugging, and refactoring. All AI-assisted content and code modifications were reviewed and validated by the authors, who retain full responsibility for the scientific content and experimental results.

\section*{Acknowledgments}
This work was supported by the European Union's Horizon Europe Research and Innovation Programme under grant agreement No. 101120732 (AUTOASSESS), the European Research Council (ERC) under grant agreement No. 864042 (AGILEFLIGHT), and the National University of Singapore.
{
    \small
    \bibliographystyle{ieeenat_fullname}
    \bibliography{bib/abbreviation_short,main}
}

\onecolumn
\setcounter{page}{1}
\appendix
{\centering
\Large
\textbf{\thetitle}\par
\vspace{0.5em}
Supplementary Material\par
\vspace{1em}
}

\section{Stage 4 Reprojection Penalty Term}
The additional bounded image-reprojection penalty near the goal introduced in Stage 4 is defined as:
\begin{equation}
r_t^{(4)}=r_t^{(1)}+r_t^{\mathrm{repr}}.
\end{equation}
Let $\mathcal G$ denote the set of goal-image pixels with valid simulator depth.
Each pixel $\mathbf u\in\mathcal G$ is back-projected into 3D, transformed into the current camera frame, and reprojected as
\begin{equation}
\begin{aligned}
\mathbf u't=\Pi\big(&R{t\leftarrow g}
[D_g(\mathbf u)\mathbf r_g(\mathbf u)]\
&+\mathbf t_{t\leftarrow g}\big),
\end{aligned}
\end{equation}
where $D_g$ is the distance from the goal camera center, $\mathbf r_g$ is the unit viewing ray, and $\Pi$ denotes calibrated camera projection.
The relative camera transform $(R_{t\leftarrow g},\mathbf t_{t\leftarrow g})$ is obtained from simulator poses.

Let $\mathcal V_t\subseteq\mathcal G$ contain points whose projections fall inside the current image.
For nonempty sets, the mean reprojection error and geometric overlap are
\begin{equation}
\begin{aligned}
e_t &= \frac{1}{|\mathcal V_t|}
\sum_{\mathbf u\in\mathcal V_t}
|\mathbf u'_t-\mathbf u|2,\
o_t &= \frac{|\mathcal V_t|}{|\mathcal G|}.
\end{aligned}
\end{equation}
We combine them into an alignment score:

\begin{equation}
A_t =
\begin{cases}
o_t \exp\left[-(e_t/\sigma)^2\right],
    & |\mathcal V_t| > 0,\\
0, & \text{otherwise},
\end{cases}
\end{equation}
where $\sigma = 5$ pixels.
The reprojection penalty is then
\begin{equation}
r_t^{\mathrm{repr}} =
\begin{cases}
-\lambda_{\mathrm{repr}}
\left(1-\frac{d_t^2}{\rho^2}\right)^2(1-A_t),
    & \substack{d_t < \rho,\\ |\mathcal G| > 0},\\
0, & \text{otherwise},
\end{cases}
\label{eq:reprojection_reward}
\end{equation}

where $d_t=|\mathbf p_t-\mathbf p_g|2$, $\rho=0.5,\mathrm{m}$, and $\lambda{\mathrm{repr}}=0.02$.
The penalty is bounded in $[-0.02,0]$, vanishes at perfect alignment, and smoothly decreases to zero at the activation boundary.
When no goal points project into the current image, the maximum distance-gated penalty is applied, while missing goal depth produces no penalty.

Geometric overlap is computed without checking occlusions in the current view.
Simulator geometry and camera poses are used only for reward computation during training and are not required by the deployed policy.

\section{Real World Deployment}
Table~\ref{tab:real_world_results} reports detailed information on each individual real world sequence.
 
\begin{table}[H]
\centering
\small
\caption{Results for the 25 real-world trials.
Errors are measured relative to the saved goal using a motion capture system.
Orientation error is the shortest full 3D rotation angle.
Time runs from policy activation to termination,
including warmup and excluding recovery.}
\label{tab:real_world_results}
\begin{tabular}{lcrrrrr}
\toprule
& & & \multicolumn{2}{c}{Initial error}
      & \multicolumn{2}{c}{Final error} \\
\cmidrule(lr){4-5}\cmidrule(lr){6-7}
Run & Outcome & Time (s)
    & Position (m) & Orientation ($^\circ$)
    & Position (cm) & Orientation ($^\circ$) \\
\midrule
0  & $\checkmark$        &  6.22 & 3.09 & 39.99 &  11.95 &  2.55 \\
1  & $\checkmark$        & 13.65 & 3.38 &  3.37 &  13.00 &  4.87 \\
2  & $\checkmark$        & 10.97 & 3.71 & 69.41 &   7.23 &  1.57 \\
3  & $\times$ (crash)    &  1.05 & 4.24 & 23.48 & 350.36 & 42.08 \\
4  & $\checkmark$        & 20.20 & 3.17 & 17.19 &   7.47 &  5.85 \\
5  & $\checkmark$        &  6.07 & 2.86 & 45.06 &  10.36 &  2.38 \\
6  & $\times$ (crash)    &  2.17 & 3.53 &  4.05 & 211.95 & 20.33 \\
7  & $\checkmark$        & 21.27 & 2.17 & 10.16 &  14.90 &  3.68 \\
8  & $\checkmark$        &  7.07 & 3.27 & 15.84 &   8.10 &  2.02 \\
9  & $\checkmark$        & 19.59 & 2.45 &  5.19 &  12.10 &  1.90 \\
10 & $\checkmark$        &  8.21 & 2.97 & 36.33 &  14.88 &  4.70 \\
11 & $\checkmark$        &  6.94 & 3.50 & 34.69 &  13.29 &  6.31 \\
12 & $\checkmark$        &  9.27 & 2.96 & 28.14 &  14.97 &  4.10 \\
13 & $\checkmark$        &  7.01 & 2.86 & 12.82 &   8.48 &  1.37 \\
14 & $\times$ (crash)    &  1.10 & 2.97 &  0.94 & 228.41 & 17.28 \\
15 & $\checkmark$        &  8.52 & 1.88 &  3.96 &  12.43 &  2.50 \\
16 & $\times$ (timeout)  & 86.52 & 2.71 & 20.83 &  38.56 &  8.32 \\
17 & $\checkmark$        & 26.66 & 2.78 &  5.32 &  14.66 &  3.74 \\
18 & $\checkmark$        &  6.66 & 3.00 &  8.82 &  10.56 &  1.19 \\
19 & $\checkmark$        &  7.71 & 3.05 &  2.98 &  12.60 &  2.66 \\
20 & $\checkmark$        &  7.33 & 2.53 & 22.74 &  14.92 &  1.48 \\
21 & $\checkmark$        & 11.06 & 2.69 & 20.04 &  12.53 &  2.99 \\
22 & $\checkmark$        &  7.63 & 3.04 & 15.91 &  13.49 &  2.42 \\
23 & $\checkmark$        &  7.56 & 3.40 & 11.37 &  11.74 &  3.33 \\
24 & $\checkmark$        &  7.28 & 3.23 & 15.15 &   8.03 &  2.92 \\
\bottomrule
\end{tabular}
\end{table}

\section{Ablations: Stage 3}
We report here the plots comparing Stage 3 trained from scratch, without the visual encoder and actor pretraining, and the full pipeline.
\begin{figure}[!htbp]
  \centering
  \includegraphics[]{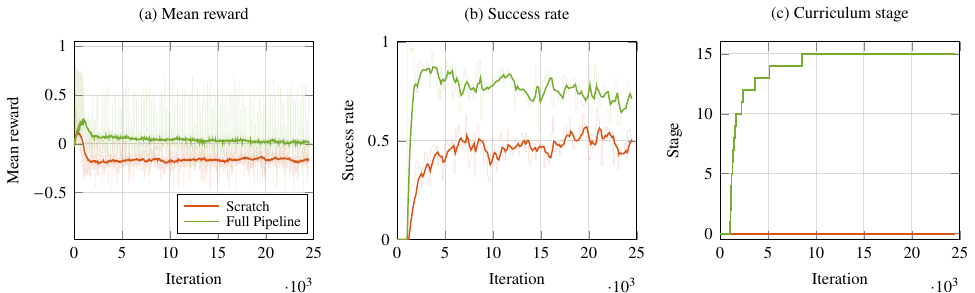}
  \caption{
Stage 3 initialization ablation. The full pipeline progresses through the curriculum, while freshly initialized run does not pass over the first curriculum level.
}
  \label{fig:stage3_loss}
\end{figure}

\section{Cross-Attention Visualization}
\label{sec:cross_attention_vis}

To better understand the learned visual representation, we visualize attention weights between current and goal image patches.
Among the inspected heads, the third head of the first cross-attention layer exhibits spatially localized responses compatible with patch correspondences.
\Cref{fig:cross_att} compares these attention maps with cosine similarities computed from frozen DINOv3 features across three HM3D validation sequences.

For each selected query patch, we visualize its attention weights over the opposite image, using Layer 1, Head 3.
The attention operates on $14\times14$ patch grids with a head dimension of 64.
For comparison, we compute cosine similarities between the frozen DINOv3 patch descriptors, before the learned adapter and attention projections.
Both maps are independently min-max normalized to $[0,1]$ for visualization.
Attention weights represent learned feature aggregation rather than explicit correspondence probabilities.

\begin{figure}[htbp]
    \centering
    \begin{subfigure}{0.9\linewidth}
        \centering
        \includegraphics[width=\linewidth,page=1]
        {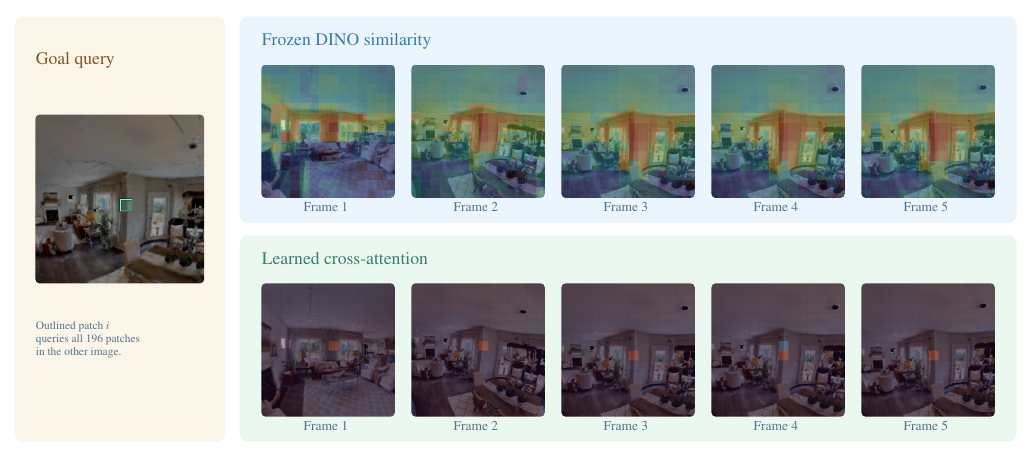}
        \caption{}
        \label{fig:cross_att_0}
    \end{subfigure}
\end{figure}

\begin{figure}[htbp]
    \ContinuedFloat
    \centering
    \begin{subfigure}{0.9\linewidth}
        \centering
        \includegraphics[width=\linewidth,page=2]
        {figures/attention_sequences.pdf}
        \caption{}
        \label{fig:cross_att_1}
    \end{subfigure}
\end{figure}

\clearpage
\begin{figure}[!ht]
    \ContinuedFloat
    \centering
    \begin{subfigure}{0.9\linewidth}
        \centering
        \includegraphics[width=\linewidth,page=3]
        {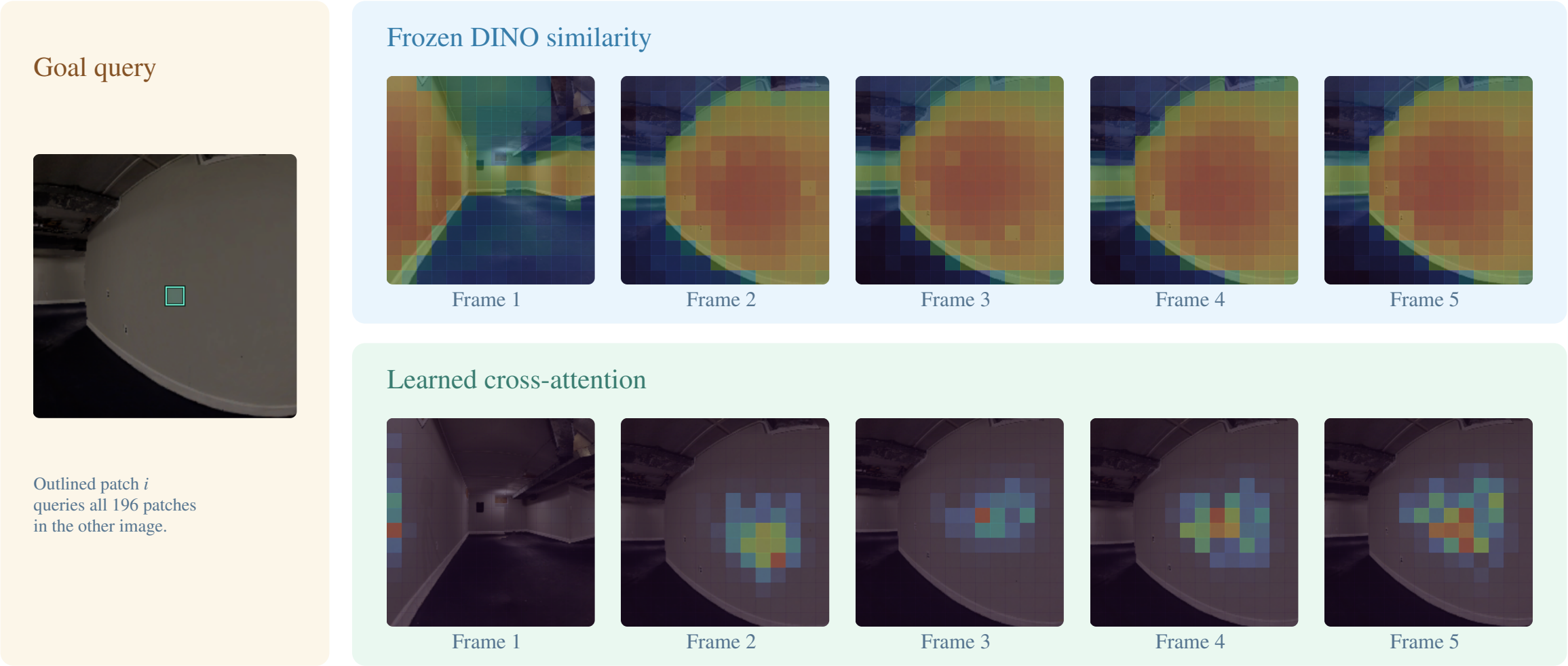}
        \caption{}
        \label{fig:cross_att_2}
    \end{subfigure}
    \caption{
        Cross-attention visualization on three HM3D validation sequences.
        (a)--(c) compare frozen DINOv3 cosine similarity with learned cross-attention from Layer 1, Head 3.
        Heatmaps are independently normalized for visualization.
    }
    \label{fig:cross_att}
\end{figure}

\section{Trajectory visualizations}
Figures~\ref{fig:real_0_traj} and~\ref{fig:hm3d_trajectories} show the first real-world flight and four HM3D validation trajectories, respectively.
Each plot shows the 3D trajectory above the goal, initial, and final images and three timestamped intermediate observations.
\newcommand{\trajectorywidth}{0.72\linewidth}
\begin{figure}[H]
  \centering
  \includegraphics[width=\trajectorywidth]{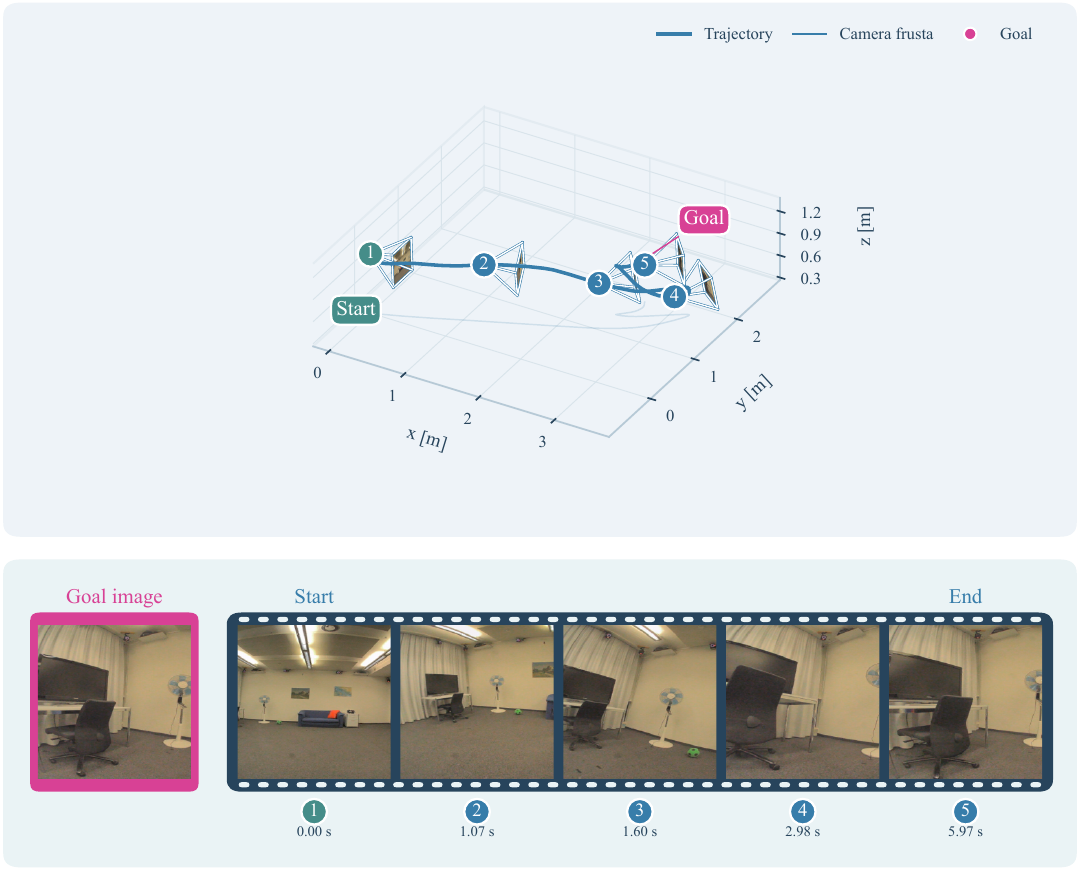}
  \caption{
  Qualitative result on a real-world flight sequence. The 3D trajectory is shown above, with the goal, initial, final, and three timestamped intermediate frames below.
}
  \label{fig:real_0_traj}
\end{figure}

\begin{figure}[p]
    \centering
    \begin{subfigure}{\trajectorywidth}
        \centering
        \includegraphics[width=\linewidth,page=1]
        {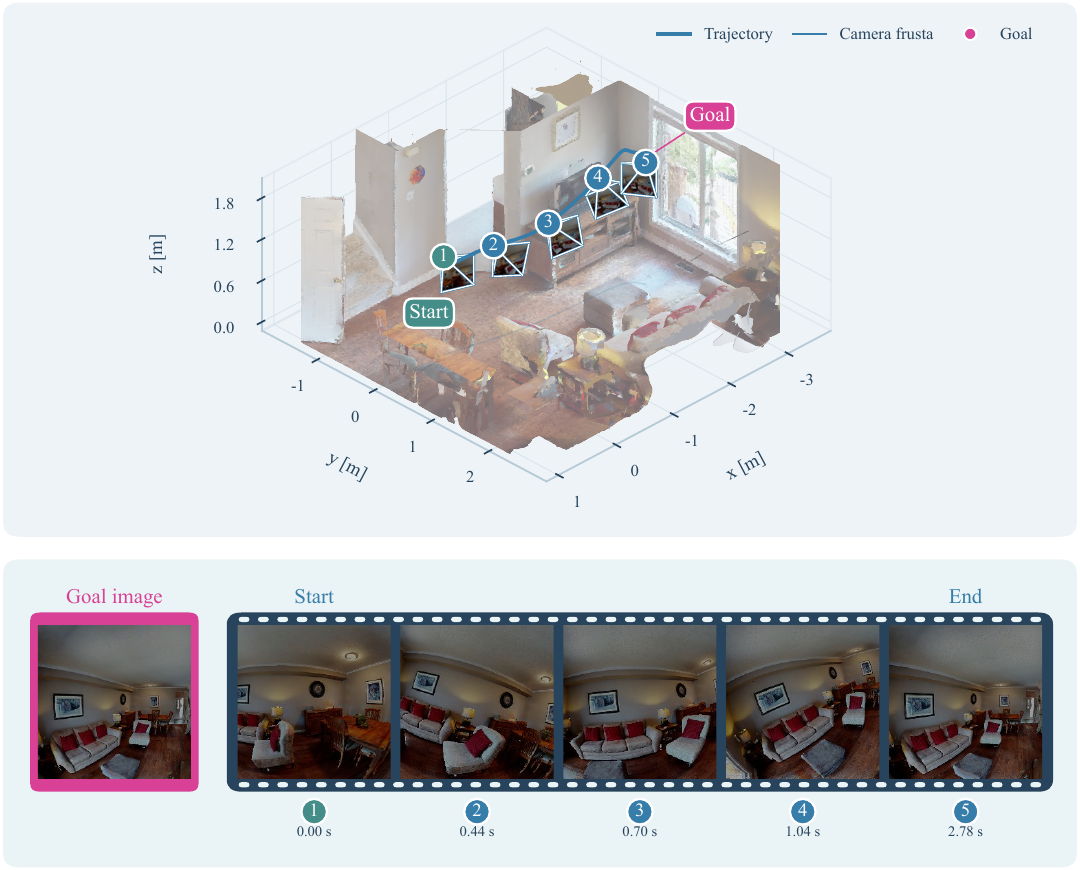}
        \caption{}
        \label{fig:hm3d_0_traj}
    \end{subfigure}\par\smallskip
    \begin{subfigure}{\trajectorywidth}
        \centering
        \includegraphics[width=\linewidth,page=2]
        {figures/hm3d_all_four_with_scenes.pdf}
        \caption{}
        \label{fig:hm3d_1_traj}
    \end{subfigure}
    \caption{Qualitative results on four HM3D validation sequences.
    Each example shows the 3D trajectory above and the goal,
    initial, final, and three timestamped intermediate frames below.}
    \label{fig:hm3d_trajectories}
\end{figure}

\begin{figure}[p]
    \ContinuedFloat
    \centering
    \begin{subfigure}{\trajectorywidth}
        \centering
        \includegraphics[width=\linewidth,page=3]
        {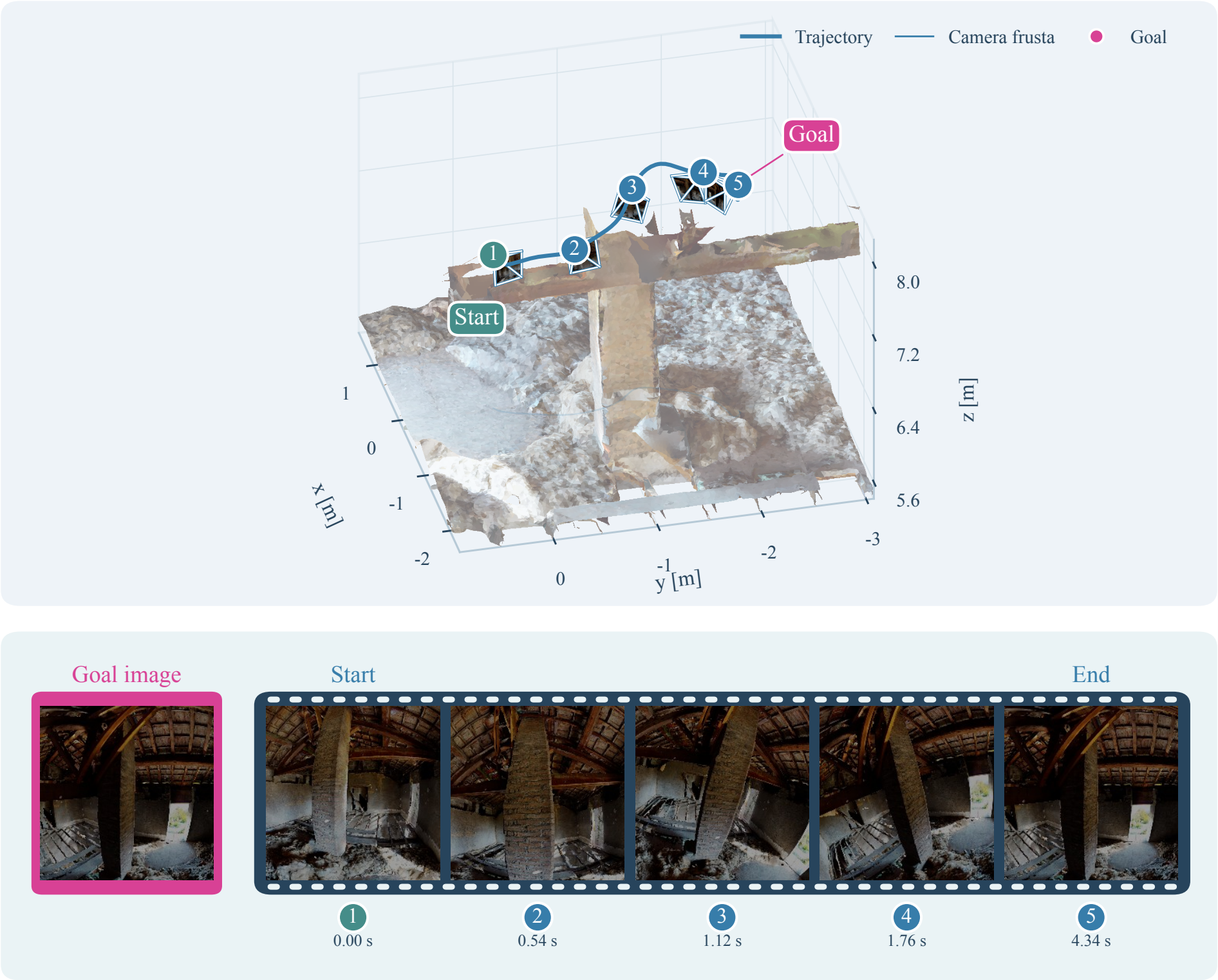}
        \caption{}
        \label{fig:hm3d_2_traj}
    \end{subfigure}\par\smallskip
    \begin{subfigure}{\trajectorywidth}
        \centering
        \includegraphics[width=\linewidth,page=4]
        {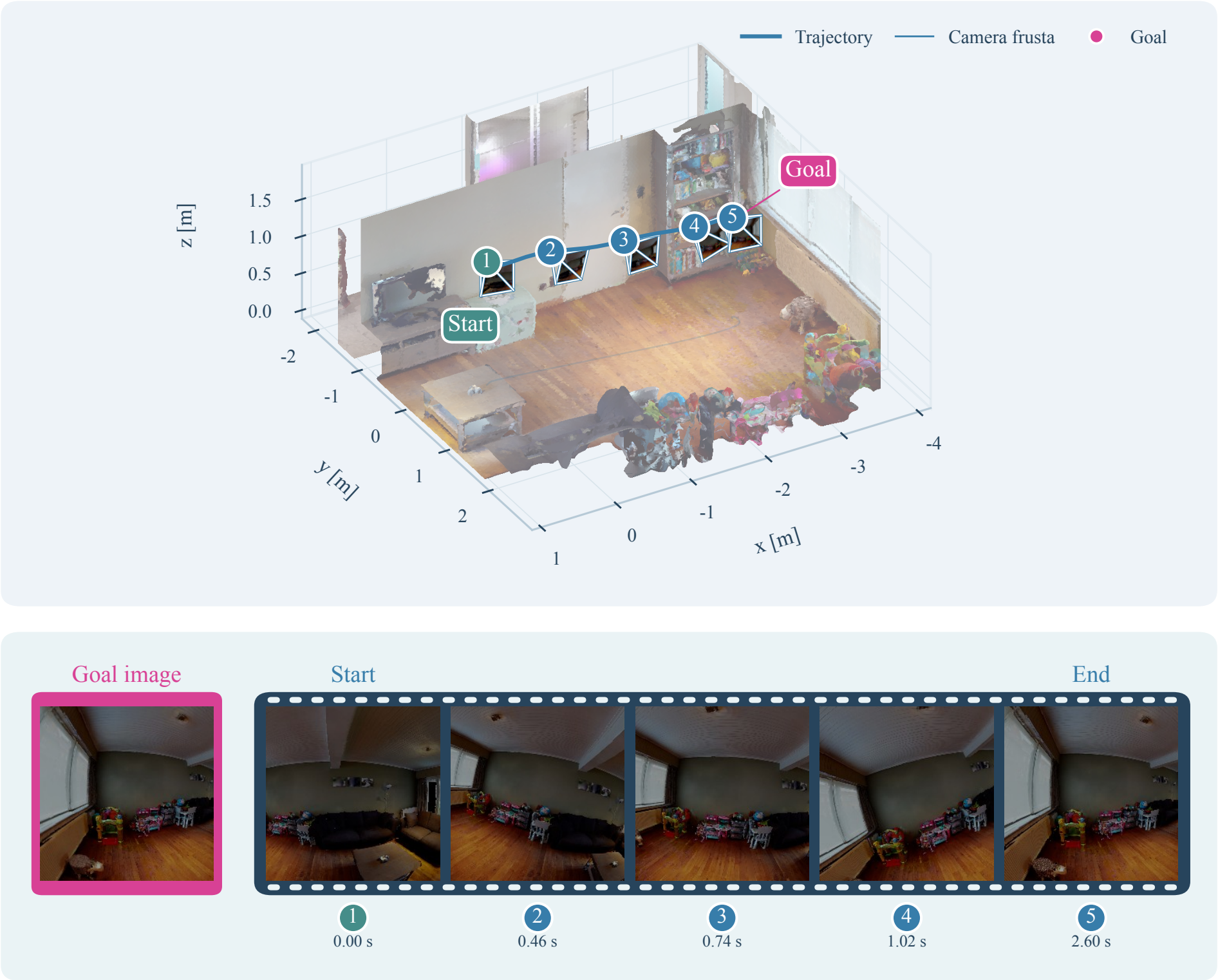}
        \caption{}
        \label{fig:hm3d_3_traj}
    \end{subfigure}
    \caption[]{HM3D validation sequences (continued).}
\end{figure}

\end{document}